\documentclass[a4paper]{article}

\usepackage{INTERSPEECH2019}
\usepackage[utf8]{inputenc}
\usepackage{bbm}
\usepackage{amssymb}
\usepackage{mystyle}
\usepackage{subcaption}

\title{Bayesian Subspace Hidden Markov Model for Acoustic Unit Discovery}
\name{Lucas Ondel, Hari Krishna Vydana, Lukáš  Burget,  Jan Černocký}
\address{Brno University of Technology}
\email{\{iondel,vydana,burget,cernocky\}@fit.vutbr.cz}

\begin{document}

\maketitle
\begin{abstract}
This work tackles the problem of learning a set of language specific acoustic units from unlabeled speech recordings given a set of labeled recordings from other languages. Our approach may be described by the following two steps procedure: first the model learns the notion of acoustic units from the labelled data and then the model uses its knowledge to find new acoustic units on the target language. We implement this process with the Bayesian Subspace Hidden Markov Model (SHMM), a model akin to the Subspace Gaussian Mixture Model (SGMM) where each low dimensional embedding represents an acoustic unit rather than just a HMM's state. The subspace is trained on 3 languages from the GlobalPhone corpus (German, Polish and Spanish) and the AUs are discovered on the TIMIT corpus. Results, measured in equivalent Phone Error Rate, show that this approach significantly outperforms previous HMM based acoustic units discovery systems and compares favorably with the Variational Auto Encoder-HMM.
\end{abstract}


\section{Introduction}

State-of-the-art Automatic Speech Recognition (ASR) systems rely upon very large amount of speech recordings paired with textual transcriptions. While this approach has proven to be very successful, it is however limited to the very few languages having enough resources to train an ASR system. Due to the cost of data collection and transcription, broadening the range of speech technologies to any language remains an unreachable objective. Parallel to the mainstream ASR, there  has been a growing interest in the paradigm of unsupervised learning of speech \cite{Glass2012}. Unsupervised speech learning attempts to use machine learning techniques to extract various information (phonetic content, speaker identity, \dots) from unlabeled recordings. While this is considerably harder than standard ASR, solving this problem would have a considerable impact on the field by reducing the amount of human labour necessary to build a full fledged ASR pipeline. It is also important to emphasize that the linguistic diversity is diminishing worldwide. Many languages are now considered endangered and risk to disappear in a near future. Affordable speech technologies could be a precious tool to help linguists and communities to document and preserve these languages.  

This work focuses on the specific task of acoustic unit discovery (AUD). Given a collection of unlabeled recordings in a specific language, the task is to learn a set of basic speech units (also called pseudo-phones) to describe the language. AUD algorithms have to solve three problems: to segment the speech, to cluster the segments into units and to infer how many units are necessary to describe the language. Several approaches have been proposed relying upon Bayesian non-parametric version of the Hidden Markov Model (HMM) \cite{Lee2012, Ondel2016, Ondel2017}. An important recent extension of this model is the VAE-HMM \cite{Ebbers2017, Glarner2018, Ondel2018} which combines the traditional HMM with Variational Auto Encoder \cite{KingmaW13}. However, most of the AUD algorithms are prone to model speaker/channel or any non-phonetic variability. To address this issue, we propose the Bayesian Subspace HMM (SHMM). The SHMM is an HMM based AUD model in which the parameters of each unit is constrained to be in the phonetic subspace of the total parameter space. This restriction forces the AUD model to focus on the phonetic content of the speech signal and to ignore irrelevant information.    

\section{Model}

\subsection{Standard Acoustic Unit Discovery}

Let $\Matrix{X} = (\Matrix{x}_1, \dots, \Matrix{x}_N)$ be the sequence of $N$ observed speech frames and $U = \{u_1, \dots, u_P\}$ be the set of $P$ acoustic units. $\Matrix{v} = (v_1, \dots, v_N), \; v_i \in U $ is a sequence of variables indicating to which unit each speech frame is associated, and $\Matrix{Z} = (\Matrix{z}_1, \dots, \Matrix{z}_N)$ are model-dependent latent variables. We consider generative models for which the complete likelihood of the data factorizes as: 
\begin{align}
    \label{eq:req1}
    p(\Matrix{X}, \Matrix{Z} | \Matrix{v}) = \prod_{n=1}^N p(\Matrix{x}_n, \Matrix{z}_n| v_n) 
\end{align}
and the likelihood of a speech frame for a given unit is member of the exponential family of distribution: 
\begin{equation}
    \label{eq:req2}
    p(\Matrix{x}_n, \Matrix{z}_n| v_n = u) = \exp \big\{\Matrix{\eta}_u^T T(\Matrix{x}_n, \Matrix{z}_n) - A(\Matrix{\eta}_u) \big\} 
\end{equation}
where $\Matrix{\eta}_u \in \mathcal{H}$ is the $D$-dimensional vector of natural parameters corresponding to one acoustic unit, $T(\Matrix{x}_n, \Matrix{z}_n)$ are the sufficient statistics and $A(\Matrix{\eta}_u)$ is the (log-)normalization constant of the density. Note that the nature of the model for the units  (HMM, GMM, Linear Dynamical Model, \dots) will depend on the value of $\Matrix{z}_n$ and the sufficient statistics $T$. In this work we consider that each unit is modeled by an HMM with a GMM for each state's emission but it can be replaced by any model satisfying Eq. \ref{eq:req1} and Eq. \ref{eq:req2}. Previous works \cite{Lee2012, Ondel2016, Glarner2018} use special cases of this model to perform the AUD. More precisely, one can understand AUD as finding a set of vectors $\Matrix{\eta}_{u_1}, \dots, \Matrix{\eta}_{u_p}$ such that the likelihood of the observation is maximized \footnote{These algorithm also learn the number of acoustic units $P$ needed to fit the data}. This search is difficult because speech recordings encode many factors other than the phonetic information (speaker identity, emotions, environment, \dots) and the AUD algorithm may maximize the likelihood while modeling non-phonetic information. 

\subsection{Subspace HMM}

\begin{figure*}
    \centering
    \begin{subfigure}{.5\textwidth}
      \centering
      \includegraphics[width=.7\linewidth]{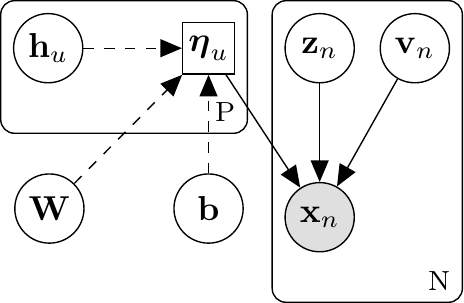}
      \caption{}
      \label{fig:sub1}
    \end{subfigure}%
    \begin{subfigure}{.5\textwidth}
      \centering
      \includegraphics[width=.65\linewidth]{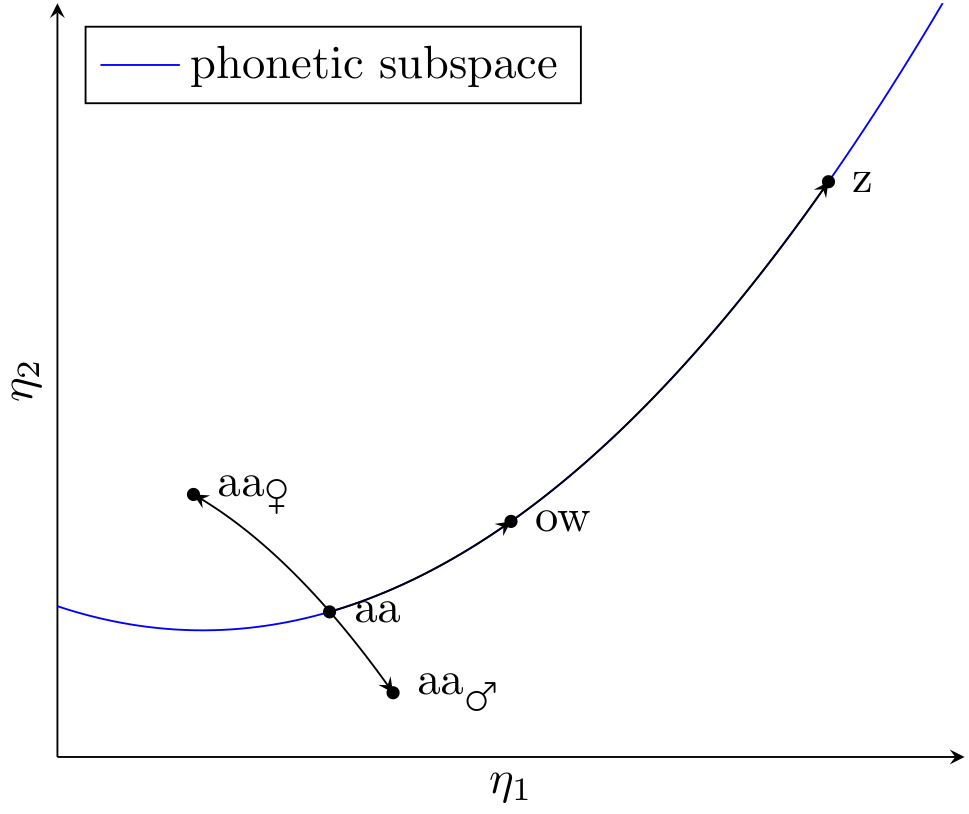}
      \caption{}
      \label{fig:sub2}
    \end{subfigure}%
    \caption{(a) Directed Acyclic Graph of a Generalized Subspace Model. Dashed lines represent deterministic relationship between variables. SHMM, JFA, SGMM are special cases of this model.  In this work each embedding $\Matrix{h}_u$ encodes the parameters of one HMM corresponding to an acoustic unit. (b) Illustration of the subspace model for acoustic units. Each point of the plane corresponds to the parameters of an acoustic unit model and the blue line represents the subspace defined by $f(\Matrix{W}^T \Matrix{h} + \Matrix{b})$. Given an acoustic unit model corresponding to the sound \emph{aa}, moving its parameters along the subspace will change the model to represent another unit/phone (\emph{ow}, \emph{z} in this example). Conversely, moving the parameters away from the phonetic subspace will push the model to capture non-phonetic information (for instance speaker gender).}
    \label{fig:shmm}
\end{figure*}
To avoid the AUD model to capture non-phonetic information, we proposed the Subspace HMM (SHMM) which constrains the parameters of the acoustic units to live in the phonetic space. This model extends the unsupervised HMM  by assuming that the phonetic information of a language is contained in a subspace of the total parameters space. Formally, it is defined as:
\begin{align}
    \label{eq:subspace}
    \Matrix{\eta}_u &= f(\Matrix{W}^T \Matrix{h}_u + \Matrix{b})
\end{align}
where $f:  \mathbb{R}^D \mapsto \mathcal{H}$ is a differentiable function. We further refine this subspace model by introducing a prior over the subspace's parameters:
\begin{align}
    W_{r,c} &\sim \mathcal{N}(0, \sigma^2_{W_{r,c}}) \\
    \Matrix{b} &\sim \mathcal{N}(\Matrix{0}, \Matrix{I}) \\
    \Matrix{h}_u &\sim \mathcal{N}(\Matrix{0}, \Matrix{I})
\end{align}
As depicted in Fig. \ref{fig:sub2}, the bases of $\Matrix{W}$ span the subspace containing the phonetic variability. Since the parameters of the acoustic units are constrained to live in a low dimensional subspace, the AUD algorithm can be seen as finding the set of embeddings $\Matrix{h}_{u_1}, \dots, \Matrix{h}_{u_P}$ which maximizes the likelihood of the observations. By constraining the search in the phonetic subspace, we therefore force the algorithm to ignore non-phonetic source of variability.

Note that Subspace Gaussian Mixture Model \cite{Povey2011}, Joint Factor Analysis \cite{Kenny2005}, Subspace Multinomial Model \cite{Kesiraju2016}, etc. are special cases of Eq. \ref{eq:subspace}. In fact, Eq. \ref{eq:subspace} is the general form of any subspace model for which the complete likelihood is a member of the exponential family of distributions. We denote Eq. \ref{eq:subspace} as the \emph{Generalized Subspace Model} (GSM) of which the Subspace HMM, like other aforementioned models, is just a special instance. The graphical representation of the (GSM) is depicted in Fig. \ref{fig:sub1}. To complete our definition of the SHMM, we need to specify the mapping $f$ from $\mathbb{R}^D$ to the natural parameters space $\mathcal{H}$. In our setting, each unit is modeled by a HMM with a 3 states left-to-right topology and each state has a GMM emissions with $K$ Gaussian components with diagonal covariance matrix. For convenience, we introduce the vector $\Matrix{\psi} = \Matrix{W}^T \Matrix{h} + \Matrix{b}$ which can be decomposed into three parts $\Matrix{\psi} = (\Matrix{\psi}_1, \Matrix{\psi}_2, \Matrix{\psi}_3)^T$. $\Matrix{\psi}_i$ is the vector of parameters (before the mapping $f$) associated with the $i$th HMM state. $\Matrix{\psi}_i$ further decomposes into $\Matrix{\psi}_i = (\Matrix{\psi}_i^{\Matrix{\pi}}, \Matrix{\psi}_{i,1}^{\Matrix{\mu}}, \dots, \Matrix{\psi}_{i,K}^{\Matrix{\mu}}, \Matrix{\psi}_{i,1}^{\Matrix{\Sigma}}, \dots, \Matrix{\psi}_{i,K}^{\Matrix{\Sigma}})^T$ where $\Matrix{\psi}_i^{\Matrix{\pi}}$ is the vector encoding the parameters of the mixture's weights and  $\Matrix{\psi}_{i,j}^{\Matrix{\mu}}$ and $\Matrix{\psi}_{i,j}^{\Matrix{\Sigma}}$ are the vectors encoding the parameters of the mean and covariance matrix of the jth Gaussian component, respectively. We set $f$ such that:
\begin{align}
    \pi_{i,j} &= \frac{\exp\{\psi^{(\Matrix{\pi})}_{i,j}\} }{1 + \sum_{k=1}^{K-1} \exp\{\psi^{(\Matrix{\pi})}_{i,k}\}} \\
    \Matrix{\mu}_{i,j} &= \Matrix{\psi}^{\Matrix{(\mu)}}_{i,j} \\
    \Matrix{\Sigma}_{i,j} &= \Diag(\exp \{ \Matrix{\psi}^{(\Matrix{\Sigma})}_{i,j} \})
\end{align}
where $\exp$ is the elementwise exponential function. One could also include the transition probabilities of the HMM but we kept them as fixed parameters in this work.

\subsection{Estimating the phonetic subspace}
Unlike previous AUD algorithms, our model requires to specify the phonetic subspace (parameterized by $\Matrix{W}$ and $\Matrix{b}$) before searching the acoustic units. This is a "chicken or egg" problem since we need the phonetic subspace to find the pseudo-phones of the language and we need to know the phones of a language to estimate the subspace. However, this problem can be alleviated by observing that many languages in the world have common phones. It is reasonable to believe that the phonetic subspace of language is well approximated by a phonetic subspace estimated from one or several other languages for which we have labeled data. Interestingly, this rationale naturally fits the Bayesian approach of the problem of AUD. Given unlabeled set of observation $\Matrix{X}^{(t)}$ in a target language $t$, previous Bayesian AUD algorithms try to estimate the inventory of (pseudo-)phones $U^{(t)}$ of the target language by estimating:
\begin{equation}
    \label{eq:std_aud}
    p(U^{(t)} | \Matrix{X}^{(t)}) = \frac{p(\Matrix{X}^{(t)}| U^{(t)})p(U^{(t)})}{p(\Matrix{X}^{(t)})}
\end{equation}
If we now assume the phonetic subspace to be estimated from the observations $\Matrix{X}^{(p)}$ of another language $p$ with known inventory of phones $U^{(p)}$, the problem can be reformulated as:
\begin{align}
    \label{eq:infprior_aud}
    p(U^{(t)} | \Matrix{X}^{(t)}, L^{(p)}, S) &= \frac{p(\Matrix{X}^{(t)}|U^{(t)}, L^{(p)}, S)p(U^{(t)} | L^{(p)}, S)}{p(\Matrix{X}^{(t)} | L^{(p)}, S)}   \\
    L^{(p)} &= \{ \Matrix{X}^{(p)}, U^{(p)} \}\\
    S &= \{ \Matrix{W}, \Matrix{b} \}
\end{align}
The term $p(U^{(t)} | L^{(p)}, S)$ may be seem as some educated/informative prior which embeds the notion of phone into the AUD algorithm. This educated prior needs to be estimated as well which leads to a two steps procedure for the SHMM AUD algorithm. First, given the labeled data of one or several languages, the prior over the acoustic units is estimated. Informally speaking, we force the model to learn "what is a phone". Second the unlabeled data of the target language is clustered into pseudo-phones given the phonetic knowledge acquired by the model during the first step.    

\subsection{Training}

The two steps of the training (learning the prior and clustering the units) are carried out by optimizing the same objective function except that when estimating the prior, the acoustic unit transcription of each utterance is known. The presence or absence of the transcription will be reflected in $p(\Matrix{v})$. When there is no transcription, $p(\Matrix{v})$ can be understood as a "pseudo-phone" loop (see \cite{Ondel2016}  for details) and when the transcription is known then $p(\Matrix{v})$ is just the inference graph used for forced alignment in a traditional HMM based ASR system. 

Since the estimation of the exact posterior of the model's parameters is intractable, we use the Variational Bayes (VB) objective function to find an approximate posterior:
\begin{align}
    \label{eq:elbo}
    \mathcal{L}[q] &= \big\langle \ln p(\Matrix{X} | \Matrix{\Xi}, \Matrix{\Theta}) \big\rangle_{q} - \KL\big( q(\Matrix{\Xi}, \Matrix{\Theta}) || p(\Matrix{\Xi}, \Matrix{\Theta}) \big) \\
    \Matrix{\Xi} &= \{ \Matrix{Z}, \Matrix{v} \} \\
    \Matrix{\Theta} &= \{ \Matrix{W}, \Matrix{b}, \Matrix{h}_{u_1}, \dots, \Matrix{h}_{u_p} \}
\end{align}
where $\langle \dots \rangle_q$ denote the expectation w.r.t. the distribution $q$ and $\KL$ is the Kullback-Leibler divergence. Eq. \ref{eq:elbo} is not tractable for arbitrary distribution $q$ we therefore consider the restricted set of distributions with the following \emph{mean-field} factorization and the given parameterization:
\begin{align}
    q(\Matrix{\Xi}, \Matrix{\Theta}) &= q(\Matrix{\Xi}; \Matrix{\phi})q(\Matrix{\Theta}; \Matrix{\zeta}) \\
    \Matrix{\zeta} &= \{ \Matrix{m}, \Matrix{\lambda} \} \\
    q(\Matrix{\Theta}; \Matrix{\zeta}) &= \mathcal{N}(\Matrix{m}, \Diag(\exp\{\Matrix{\lambda}\})
\end{align}
The parameters $\Matrix{\phi}$ of the variational posterior over $\Matrix{\Xi}$ will depend on the type of the model of the acoustic unit. For the case of an HMM, this is the probability to be in a particular state given the sequence of observations. Under these restrictions the optimization reduces to:
\begin{align}
    \Matrix{\phi}^*, \Matrix{\zeta}^* = \argmax_{\Matrix{\phi}, \Matrix{\zeta}} \mathcal{L}(\Matrix{\phi}, \Matrix{\zeta})
\end{align}
Since we assume each unit to be modeled by an HMM, $\Matrix{\phi}^*$ has an analytical solution which can be efficiently calculated using the \emph{forward-backward} algorithm \cite{Rabiner1989}. $\Matrix{\zeta}^*$ has no analytical solution but can be found through a stochastic gradient ascent scheme. Noting that $\nabla_{\Matrix{\phi}}\mathcal{L}_{\Matrix{\zeta}}(\Matrix{\phi}^*) = \Matrix{0}$ we have: 
\begin{align}
    \nabla_{\Matrix{\zeta}} \mathcal{L}(\Matrix{\phi}^*, \Matrix{\zeta}) &= \nabla_{\Matrix{\zeta}} \mathcal{L}_{\Matrix{\phi}^*}(\Matrix{\zeta})  + \nabla_{\Matrix{\zeta}} \Matrix{\phi}^* \nabla_{\Matrix{\phi}}\mathcal{L}_{\Matrix{\zeta}}(\Matrix{\phi}^*) \\
    &= \nabla_{\Matrix{\zeta}} \mathcal{L}_{\Matrix{\phi}^*}(\Matrix{\zeta})
\end{align}
Finally, we approximate $\nabla_{\Matrix{\zeta}} \mathcal{L}(\Matrix{\phi}^*, \Matrix{\zeta}) \approx \nabla_{\Matrix{\zeta}} \mathcal{L}'(\Matrix{\phi}^*, \Matrix{\zeta})$ by using the so called \emph{re-parameterization} trick introduced in \cite{KingmaW13}:
\begin{align}
    \Matrix{\epsilon}_l &\sim \mathcal{N}(\Matrix{0}, \Matrix{I}) \\
    \Matrix{\Theta}_l &= \Matrix{m} + \Diag(\exp\{\frac{\Matrix{\lambda}}{2}\}) \Matrix{\epsilon}_l \\
    \mathcal{L}(\Matrix{\phi}, \Matrix{\zeta}) &\approx \frac{1}{L} \sum_{l=1}^L \ln p(\Matrix{X} | \Matrix{\Xi}, \Matrix{\Theta}_l) \\ 
    & \quad - \KL\big( q(\Matrix{\Xi}, \Matrix{\Theta}) || p(\Matrix{\Xi}, \Matrix{\Theta}) \big) = \mathcal{L}'(\Matrix{\phi}, \Matrix{\zeta})
\end{align}
In practice we use the ADAM optimizer \cite{KingmaB14} to update $\Matrix{\zeta}$ and we use $L = 10$ samples to compute the empirical expectation. The parameters $\Matrix{\phi}$ are re-estimated every $1000$ updates of $\Matrix{\zeta}$.

\section{Experiments}

\subsection{Data, Features and Metrics}

We conducted our experiments with the TIMIT \cite{timit} database and 3 languages from the GlobalPhone corpus \cite{globalphone}: German (GE), Polish (PO) and Spanish (SP). For each of the three GlobalPhone languages, we kept only 3000 randomly selected utterances. We used two sets of features: (i) the MFCC features concatenated with their first and second derivatives (ii) the Multi-Lingual bottleNeck (MBN) features trained on 17 Babel's languages \cite{Fer2017}. The set of languages used to train the MBN features does not include English, German, Polish or Spanish. Both set of features were extracted at a rate of 100 Hz. For the case the MBN features, the audio signal was down-sampled to 8kHz. 

We evaluated the different AUD algorithms in terms of phonetic segmentation and equivalent Phone Error Rate (eq. PER) (\cite{KamperJG17, Ebbers2017}). For the phonetic segmentation we used the standard Recall, Precision and F-score measured against the timing provided in the TIMIT database with the 61 original phones. We tolerated boundary shifted by +- 2 frames (20 milliseconds). To compute the eq. PER, we mapped each acoustic unit to one of the 61 phones it overlaps the most with. Then, we reduced the reference transcription and proposed transcription to the 39 phone set \cite{kaifulee1989speekerindependent} and computed the PER.   

\begin{table*}[htb]
    \centering
    \scalebox{0.7}{
    \begin{tabular}{|c|c|c|c|c|c|c|}
            \hline 
            Model & Features & Prior Language & Recall & Precision & F-score &  eq. PER \\
            \hline 
            HMM \cite{Ebbers2017} & MFCC + $\Delta$ + $\Delta \Delta$ & None & -  & - & - &  65.4 \\
            VAE-HMM \cite{Ebbers2017}  & MFCC + $\Delta$ + $\Delta \Delta$ & None  & - & - & - & 58.9 \\ 
            VAE-BHMM \cite{Glarner2018}  & log-mel FBANK + $\Delta$ + $\Delta \Delta$ & None  & - & - & - & 56.57 \\ 
            HMM & MFCC + $\Delta$ + $\Delta \Delta$ & None & 66.47 & 57.81 & 61.84 & 64.92  \\
            HMM & MBN & None & 63.98 & 54.21 & 58.69 & 68.25  \\
            \hline
            SHMM & MFCC + $\Delta$ + $\Delta \Delta$ & GE & \textbf{75.74} & \textbf{78.98} & \textbf{77.32} & 58.89 \\
            SHMM & MFCC + $\Delta$ + $\Delta \Delta$ & GE+PO & 73.94 & 74.47 & 74.20 & 58.23 \\
            SHMM & MFCC + $\Delta$ + $\Delta \Delta$ & GE+PO+SP & 75.03 & 74.00 & 74.51 & 56.91 \\
            \hline
            SHMM & MBN & GE & 56.57 & 69.34 & 62.31 & 55.14 \\
            SHMM & MBN & GE+PO & 59.18 & 69.12 & 63.76 & 54.1 \\
            SHMM & MBN & GE+PO+SP & 60.89 & 68.41 & 64.43 & \textbf{49.2} \\
            \hline
        \end{tabular}}
    \caption{Comparison of the SHMM against other AUD models in terms of phonetic segmentation (Recall, Precision, F-score) and equivalent Phone Error Rate (\%).}
    \label{tab:main_results}
\end{table*}

\subsection{Estimating the phonetic subspace}

First, we ran a controlled experiment to assess whether the SHMM is able to properly learn the phonetic subspace of a language. In this experiment, we used the MBN features and each HMM state had 8 Gaussian components. First, we trained a Bayesian HMM phone recognizer on the 48 phone set with a flat phonotactic language model on the traditional TIMIT training set (no SA* utterances) and decoded on the test set mapping the phones to the 39 phone set. This phone recognizer achieved 36.4 \% Phone Error Rate (PER). This number is very high since we have removed crucial elements of the traditional ASR pipeline (language model, context-dependent phones, \dots) in order to evaluate the quality the acoustic model. For comparison, we trained a monophone system with a flat phonotactic language model using the Kaldi toolkit \cite{Povey2011} which yielded 37.3 \% PER. We then trained an SHMM based phone recognizer with varying subspace dimension using the same training and testing setup as the baseline HMM. We used the baseline model to provide the first estimate of $\Matrix{\phi}$ which we modified so that all the Gaussian components within a state  have equal  responsibility. We pre-trained the subspace for 15000 updates before updating $\Matrix{\phi}$ then we re-estimated $\Matrix{\phi}$ after every 1000 updates of $\Matrix{\zeta}$ for 30 iterations. 
\begin{figure}[ht]
    \centering
    \includegraphics[scale=.25]{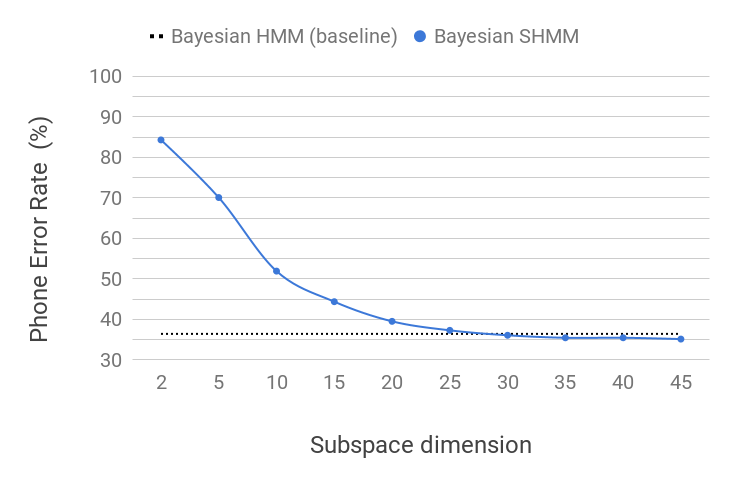}
    \caption{PER of the SHMM for varying subspace dimension.}
    \label{fig:per_vs_sdim}
\end{figure}
Results, shown in Fig. \ref{fig:per_vs_sdim}, indicate that the SHMM is perfectly able to learn the phonetic subspace of a language by compressing the 3861-dimensional \footnote{3 states $\times (8 \text{ Gaussian } \times 2 \times 80 + 7)$. 80 is the features dimension, $2$ accounts for the mean and the diagonal of the covariance matrix and 7 is the dimension of the per-state mixture weights.} parameter space to a subspace as small as 30 dimensions and yet achieving the same PER as the HMM baseline.

\subsection{Acoustic Unit Discovery}

We now consider the case of unsupervised learning of speech where English is assumed to be a low-resourced language. In this setup, we use the complete TIMIT set (training, development and testing set including the SA* utterances) as the corpus from which to extract acoustic units. In this experiment, all the HMM/SHMM have 4 Gaussian components per state. Our baseline is the HMM based AUD system described in \cite{Ondel2016} and the VAE-(B)HMM based AUD system proposed in \cite{Ebbers2017, Glarner2018}. We compare these baselines with 3 SHMM based AUD models for which the posterior of the phonetic subspace $q(\Matrix{W}, \Matrix{b})$ was estimated using: (i) German language (ii) German and Polish languages (iii) German, Polish and Spanish languages. For each case the phonetic subspace had 35, 70 and 100 dimensions respectively. 
Note that the choice of the languages and the order of combination was arbitrary and it is likely that choosing languages closely related to the target language would be beneficial. We considered all the phones of all the languages to be unique and didn't merge them while estimating the subspace. The posteriors of the embeddings $q(\Matrix{h}_u)$ corresponding to the German, Polish and Spanish phones were discarded before the AUD clustering. 

The results are presented in Table \ref{tab:main_results} and differ significantly depending on the input features. The SHMM always benefits from learning the phonetic subspace in terms of eq. PER. Interestingly, the baseline HMM fails to benefit from the MBN features as it underperforms compared to the HMM trained on MFCC features. The SHMM, thanks to its subspace, learns from other languages to fully exploit the discriminatively trained features. Regarding the segmentation evaluation, the SHMM better segments the speech compared to the simple HMM. However, we observe that using more than one language does not necessarily improves the segmentation. Also, contrary to the eq. PER, the MBN features does not seem to be ideal to get accurate segmentation. 

Finally, we tried to label the TIMIT corpus with a HMM phone-recognizer (MBN features) trained on German, German and Polish and German, Polish and Spanish and we interpreted the output phones as acoustic units. For these 3 models the eq. PER was 61.22 \% (GE), 66.47 \% (GE+PO) and 71.96 \% (GE+PO+SP). Contrary to the SHMM, this naive approach does not benefit from having more languages.

\section{Conclusions}
We proposed a new model for AUD: the Subspace HMM. Unlike other AUD models the SHMM is trained in a supervised fashion on one or several languages to learn the notion of "phone". This phonetic knowledge is encoded into a non-linear subspace of the total parameter space. Then, the SHMM searches a set of of acoustic units in this subspace which maximizes the likelihood of the observations of the target language. The SHMM outperforms the HMM based AUD and is competitive with the VAE-HMM. When using discriminatively trained features, the SHMM achieves 49.2 \% equivalent PER on TIMIT whithout any supervision in the target language. 
\section{Acknowledgements}

The work was supported by Czech National Science Foundation (GACR) project "NEUREM3" No. 19-26934X, Czech Ministry of Interior project No. VI20152020025 "DRAPAK", and Czech Ministry of Education, Youth and Sports from the National Programme of Sustainability (NPU II) project "IT4Innovations excellence in science - LQ1602". This work was also supported by by the Office of the Director of National Intelligence (ODNI), Intelligence Advanced Research Projects Activity (IARPA) MATERIAL program, via Air Force Research Laboratory (AFRL) contract \# FA8650-17-C-9118. The views and conclusions contained herein are those of the authors and should not be interpreted as necessarily representing the official policies, either expressed or implied, of ODNI, IARPA, AFRL or the U.S. Government. 

The authors would like to thank Kateřina Žmolíková for her corrections and insightful comments.

\bibliographystyle{IEEEtran}

\bibliography{mybib}


\end{document}